# Multi-Layer Competitive-Cooperative Framework for Performance Enhancement of Differential Evolution

**Sheng Xin Zhang, Li Ming Zheng, Kit Sang Tang, Shao Yong Zheng, Wing Shing Chan**

# Multi-Layer Competitive-Cooperative Framework for Performance Enhancement of Differential Evolution

Sheng Xin Zhang [a, b], Li Ming Zheng [c], Kit Sang Tang [b], Shao Yong Zheng [a], Wing Shing Chan [b]

[a] School of Electronics and Information Technology, Sun Yat-sen University, Guangzhou, 510006, China
[b] Department of Electronic Engineering, City University of Hong Kong, Kowloon, Hong Kong
[c] Department of Electronic Engineering, School of Information Science and Technology, Jinan University, Guangzhou, 510632, China

**Abstract**

Differential Evolution (DE) is recognized as one of the most powerful optimizers in the evolutionary algorithm (EA) family. Many DE variants were proposed in recent years, but significant differences in performances between them are hardly observed. Therefore, this paper suggests a multi-layer competitive-cooperative (MLCC) framework to facilitate the competition and cooperation of multiple DEs, which in turns, achieve a significant performance improvement. Unlike other multi-method strategies which adopt a multi-population based structure, with individuals only evolving in their corresponding subpopulations, MLCC implements a parallel structure with the entire population simultaneously monitored by multiple DEs assigned to their corresponding layers. An individual can store, utilize and update its evolution information in different layers based on an individual preference based layer selecting (IPLS) mechanism and a computational resource allocation bias (RAB) mechanism. In IPLS, individuals connect to only one favorite layer. While in RAB, high-quality solutions are evolved by considering all the layers. Thus DEs associated in the layers work in a competitive and cooperative manner. The proposed MLCC framework has been implemented on several highly competitive DEs. Experimental studies show that the MLCC variants significantly outperform the baseline DEs as well as several state-of-the-art and up-to-date DEs on CEC benchmark functions.

## 1. Introduction

Differential evolution (DE) [30] is well known for its efficiency in solving various continuous optimization problems [7, 8, 24]. DE has been widely explored over the past two decades, and consequently many advanced DE variants have been proposed. Recently, several competitive DEs, including CoBiDE (DE with covariance matrix learning and bimodal distribution parameter setting) [39], SHADE (DE with success-history based parameter adaptation) [33], MPEDE (DE with multi-population ensemble) [44] and IDE (DE with individual-dependent mechanism) [36] have been designed. However, when compared with each other, there were not many differences in their performances. It still remains a challenge to construct a DE algorithm that can significantly outperform all of these up-to-date DEs.

On the other hand, since the proposal of the AMALGAM-SO (A multi-algorithm genetically adaptive method for single objective optimization) [37] algorithm, research [1, 13, 25, 28, 31, 44, 50] on combining multiple operators or multiple evolutionary algorithms (EAs) have been a hot topic in the EA community. These methods usually employ a multi-population structure, which divides the entire population into several subpopulations. However, when the population size of each constituent optimizer is large, the convergence of the hybrid algorithm may decrease significantly. The clustering of individuals would slow down and function evaluations more likely spent on random explorative moves [26]. Moreover, it may be difficult to incorporate some complex variants and there is still uncertainty on how to take advantage of these optimizers simultaneously.

Recognizing the distinct merits of different DE designs and the difficulties in managing multiple DEs under a multi-population structure, we aim to propose a flexible framework that is able to combine multiple DEs efficiently and achieve a significant improvement of performance. A multi-layer competitive-cooperative (MLCC) framework is hence developed in this paper. In MLCC, a single population is maintained, while, by deploying individuals to operational layers, it facilitates competition and cooperation amongst the employed DEs. Features of MLCC are highlighted as follows:

(1) Different from existing multi-population based hybrid methods, MLCC introduces a parallel multi-layer structure with each layer associated with one adaptive DE optimizer. This parallel structure is expected to i) eliminate the significant increase in population size as observed in existing multi-population based structures; and ii) preserve the original designs of the constituent optimizers, providing high flexibility to incorporate complex DE variants.

(2) Competition in MLCC is designed to efficiently distribute computation resources. This is accomplished by the individual preference-based layer selecting (IPLS) mechanism, that allows each individual to connect to its favorite layer. IPLS differs from existing methods [21, 25, 28, 37, 42-44] in three aspects: i) each layer in MLCC has access to the entire population. Although some individuals (i.e. the target vectors) may be processed by a specific layer at some time, individuals for mutation can be selected from the entire population; ii) each individual can store, use and update its evolution information in multiple layers.

This facilitates the incorporation of self-adaptive DEs [4, 22, 39], which have to evolve individual specified strategies or parameters; and iii) the entire population is monitored by multiple layers to help each optimizer make decisions based on the current evolution stage.

(3) Cooperation in MLCC takes advantages of all the constituent optimizers simultaneously, to allow them to collaborate closely. This is realized using the resource allocation bias (RAB) mechanism. In RAB, some high-quality solutions are allowed to generate multiple trial vectors by using all the layers while the inferior solutions only produce one trial vector. RAB is designed based on the following considerations: i) simultaneous consideration of all the layers for superior individuals can provide multiple directions for evolution; ii) the layers in MLCC usually have complementary properties. Evolving elitism solutions by these layers simultaneously is less likely to suffer from a local optimum but instead enhances the exploitation capability of the algorithm; iii) different from canonical DE [30] and existing DEs, RAB allocates more resources to superior solutions. As a result, the evolution can put more efforts onto promising searching directions, which may be beneficial to the entire population later; and iv) the same as canonical DE, inferior solutions in RAB can still generate their offspring. This ensures that the chances of inferior solutions can compete with the superior ones thus keeping the exploratory capability of DE.

The effectiveness of the proposed MLCC framework and its components, i.e. IPLS and RAB, have been verified through extensive experiments conducted using 30 benchmark functions derived from the 2014 IEEE Congress on Evolutionary Computation (IEEE CEC2014) [18]. Numerical results show that MLCC significantly improves the performance of the baseline DEs. Moreover, the resulting MLCC variant significantly outperforms state-of-the-art and up-to-date DEs.

The remainder of this paper is organized as follows: Sec. 2 briefly reviews the related works. Sec. 3 describes the proposed MLCC framework and its implementation details. Sec. 4 presents the experiments and discussions. Finally, Sec. 5 concludes this paper.

## 2. Background and Related Works

### 2.1 Basics of DE

DE is a population-based stochastic search method for continuous real parameter optimization problems. Given a $D$-dimensional minimization problem, DE begins with a population of $NP$ individuals,

$P_0 = \left\{ \vec{x}_{i,0} = (x_{i,1,0}, x_{i,2,0}, \cdots, x_{i,D,0}), i \in \Phi \equiv \{1, 2, \cdots NP\} \right\}$ randomly sampled from the searching space. Afterwards, at each generation $G$, three operations: mutation, crossover and selection are performed. They are briefly introduced in the following.

*Mutation:* In mutation, a mutant vector $\vec{v}_{i,G}$ corresponding to each target vector $\vec{x}_{i,G}$ is generated by combining a base vector with one or more difference vectors. Frequently used mutation strategies include:

1) DE/rand/1

$$\vec{v}_{i,G} = \vec{x}_{r_1,G} + F(\vec{x}_{r_2,G} - \vec{x}_{r_3,G}) \tag{1}$$

2) DE/best/1

$$\vec{v}_{i,G} = \vec{x}_{best,G} + F(\vec{x}_{r_1,G} - \vec{x}_{r_2,G}) \tag{2}$$

3) DE/rand/2

$$\vec{v}_{i,G} = \vec{x}_{r_1,G} + F(\vec{x}_{r_2,G} - \vec{x}_{r_3,G}) + F(\vec{x}_{r_4,G} - \vec{x}_{r_5,G}) \tag{3}$$

4) DE/best/2

$$\vec{v}_{i,G} = \vec{x}_{best,G} + F(\vec{x}_{r_1,G} - \vec{x}_{r_2,G}) + F(\vec{x}_{r_3,G} - \vec{x}_{r_4,G}) \tag{4}$$

5) DE/current-to-best/1

$$\vec{v}_{i,G} = \vec{x}_{i,G} + F(\vec{x}_{best,G} - \vec{x}_{i,G}) + F(\vec{x}_{r_1,G} - \vec{x}_{r_2,G}) \tag{5}$$

where $\vec{x}_{best,G}$ denotes the best vector in the current population $P_G$, $r_m \in \Phi \setminus \{i\}$ with $m = 1,2,\cdots 5$, are distinct integers and $F$ is a user-specified mutation control parameter within (0, 1].

*Crossover:* After mutation, crossover is performed on each mutant vector $\vec{v}_{i,G}$ and its corresponding target vector $\vec{x}_{i,G}$ to generate a trial vector $\vec{u}_{i,G}$. The classic binomial crossover is formulated as follows.

$$u_{i,j,G} = \begin{cases} v_{i,j,G} & \text{if } rand_j(0,1) < CR \text{ or } j = j_{rand} \\ x_{i,j,G} & \text{otherwise} \end{cases} \tag{6}$$

where $rand_j(0,1)$ is a uniform random number within (0, 1), $j_{rand}$ is a randomly generated integer from [1, $D$], and $CR$ is a user-defined crossover control parameter within [0,1].

*Selection*: Selection is to determine the better vector between $\vec{u}_{i,G}$ and $\vec{x}_{i,G}$ which will survive in the next generation, based on their fitness values $f(\cdot)$.

$$\vec{x}_{i,G+1} = \begin{cases} \vec{u}_{i,G} & \text{if } f(\vec{u}_{i,G}) \leq f(\vec{x}_{i,G}) \\ \vec{x}_{i,G} & \text{otherwise} \end{cases} \tag{7}$$

## 2.2 Advanced DE variants

Since its advent, DE has attracted a lot of attention and many DE variants [8, 24] have been proposed with different characteristics. Among them, self-adaptive [4, 22, 39] and adaptive [29, 33, 46] DEs exhibit encouraging performance [2].

*Self-adaptive DEs:* Self-adaptive DEs [4, 22, 39] allow adjustments of strategy and/or parameter settings in each individual during evolution. The self-adaptive DE (jDE) [4] encodes control parameters $F$ and $CR$ into each individual and makes them self-adaptive during the evolution. The Parameters and Mutation

Strategies Ensemble DE (EPSDE) [22] assigns mutation strategies and control parameters to individuals from a preset pool. The Covariance matrix Learning and Bimodal Distribution Parameter Setting Based DE (CoBiDE) [39] introduces a self-adaptive bimodal parameter sample scheme.

*Adaptive DEs:* Adaptive DEs [29, 33, 46] usually collect population-wise success experience from previous generations and then use it as a guideline for later evolution. The Strategy Adaptive DE (SaDE) [29] dynamically determines the selecting probabilities of four mutation strategies according to their previous performances. The Adaptive DE (JADE) [46] introduces a new "current-to-pbest/1" mutation strategy and a success-based parameter adaptation mechanism. Due to its impressive performance, JADE was later modified, giving birth to many variants, such as Success History Based Adaptive DE (SHADE) [33], Linear Population Size Reduction Based SHADE (L-SHADE) [34], Collective Information Powered DE (CIPDE) [47] and Selective Candidate with Similariy Selection Rule (SCSS) based variants [45].

In addition, with the support of various mutation strategies, plentiful multiple strategies based DEs have also been proposed. The Composite DE (CoDE) [38] adopts three mutation strategies with different pairs of *F* and *CR* to generate offspring. The Multi-population Based Ensemble of Multiple Strategies DE (MPEDE) [44], Multiple Subpopulations Based Adaptive DE (MPADE) [6] and Individual-dependent DE (IDE) [36] assign different mutation strategies to different subpopulations. Apart from single-objective optimization, mutation strategy selection has also been extended to multi-objective optimization [19, 20].

Besides, the mutation and crossover operations of DE have also been improved using various mechanisms, such as ranking based mutation [12], two-step subpopulation based mutation [48], eigenvector based crossover [10], hybrid linkage crossover [5] and orthogonal crossover [40], to name a few.

## 2.3 Multi-method Search

According to the No Free Lunch Theorem (NFL) [41], no single algorithm or setting can perform the best for all kinds of problems. For this reason, many researchers have put much efforts into the ensemble of multiple operators or multiple EAs into their algorithms to confront different challenges in different evolutionary stages, which in turn improve the overall performance.

Vrugt et al. [37] merged multiple EAs, including Genetic Algorithm (GA) [14], Covariance Matrix Adaptation Evolution Strategy (CMA-ES) [15], Particle Swarm Optimizer (PSO) [49] and DE together to formulate the Multi-algorithm Genetically Adaptive Method for Single Objective Optimization (AMALGAM-SO) to promote efficient searches. AMALGAM-SO dynamically adjusts subpopulation size for each constituent algorithm according to its previously achieved performance. Peng et al. [25] proposed a population-based algorithm portfolio (PAP) scheme, in which each constituent optimizer is given a preset time budget to run, while different optimizers are allowed to interact with each other based on a migration strategy. Gong et al. [13] proposed a cheap surrogate model to estimate the densities of multiple candidates produced by multiple operators and then select the one with maximum density as offspring. Iacca et al. [16] suggested the Multi-strategy Coevolving Aging Particle Optimizer (MS-CAP), combining the advantages of aging based PSO and multiple strategies based DE. In [23], Noman et al. proposed an adaptive local search method to improve the performance of classic DE. In [27], Piotrowski et al. introduced a memetic DE by incorporating a local search algorithm into an adaptive DE. In [17], Kämpf et al. proposed a hybrid algorithm which combined CMA-ES with a hybrid DE. In [21], Li et al. designed a Hybrid DE (HDE) framework to perform two DEs alternatively during the evolution process. In HDE, only one of the DEs is activated in each generation. If the DE that is running is regarded as inefficient, it will be replaced by the other one.

## 3. Proposed MLCC Framework

### 3.1 Motivation

From the literature reviews presented in Sec. 2, it can be observed that: 1) Existing multi-method search [25, 37, 44] commonly divides the entire population into several subpopulations where each subpopulation evolves with an associated method. This approach may result in two drawbacks. Firstly, as recommended in many studies, eg. [30], the subpopulation size needs to be large enough to ensure a promising performance. Therefore, it may result in a large population size, which leads to a deterioration in exploitation capability of the constructed algorithms. Secondly, integrating multiple DE variants under the multi-method searching framework may not be easy. For example, the multi-method search in [25, 37, 44] had to be modified empirically for new DE variants. Also, some complex variants, such as IDE [36] which is already a multipopulation based algorithm, are hard to integrate. Furthermore, it still remains a mystery on how to take advantage of all employed methods to achieve more promising search directions; 2) Since the introduction of SaDE (Strategy adaptive DE) [29] algorithm, the concept of adaptation has been widely adopted in designing DE variants, usually realized with multiple mutation strategies [16, 19, 22, 44], multiple crossover strategies [8, 23], and the adjustments of control parameters [33, 39, 46]. Regarding the parameter mechanism, several self-adaptive [4, 22, 39] and adaptive [29, 33, 46] methods have been proposed. However, due to the fact that a single mechanism may consistently generate parameters with fixed characteristics, it may hinder the capability of an algorithm to seek out better parameters; 3) Existing DEs evolve each individual with equal amount of efforts, despite their potentials.

With the above considerations, this paper proposes a parallel multi-layer structure based competitive-cooperative (MLCC) DE framework, empowered by two mechanisms, namely the individual preference-based layer selecting (IPLS) mechanism and the computational resource allocation bias (RAB) mechanism.

### 3.2 Competitive IPLS mechanism

Figure 1 depicts the proposed parallel multi-layer structure and IPLS mechanism, in which $M$ different methods are associated with $M$ layers $L_m$ ($m = 1, 2, \ldots, M$) and there are $NP$ individuals $\{\vec{x}_{i,G}, i \in \{1, 2, \cdots NP\}\}$ in the population.

The multi-layer structure is designed as follows. In every generation, the entire population is monitored simultaneously by multiple optimizers assigned in their corresponding layers. Each target individual is assigned to a specific layer at a particular time based on its preference (for example, individual 2 is assigned to $L_1$ in Fig. 1(a)). However, it is still possible to select any individual in the entire population for the mutation process in a layer.

Preference of the target individuals is determined by the IPLS mechanism (**Algorithm 1**), described as follows.

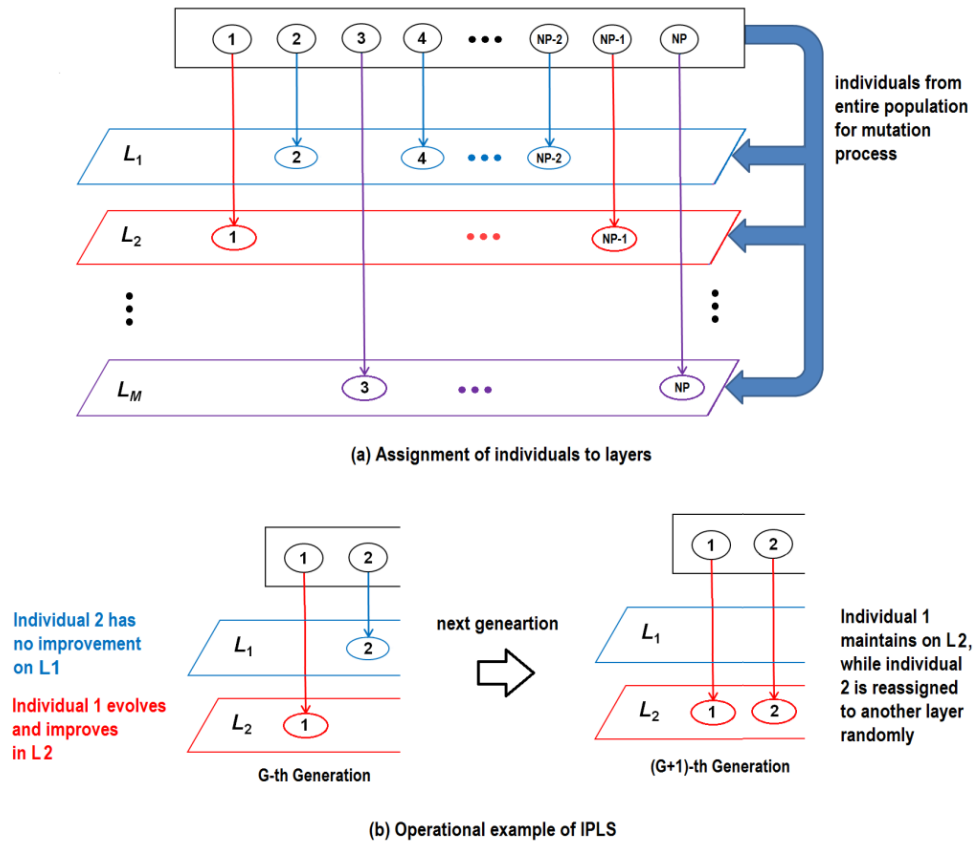

**(a) Assignment of individuals to layers**

**(b) Operational example of IPLS**

Fig.1 Illustration of the parallel multi-layer structure and IPLS mechanism. (a) Each individual connects to a layer, denoted by the solid circle while each layer can select any individual from the entire population for the mutation process of the DE employed. (b) Update of layer assignment by IPLS.

At generation $G = 0$, individual preferences $\{IP_{i,G}, i \in \{1, 2, \cdots NP\}\}$ are randomly initialized (line 1 in Algorithm 1). Afterwards, for each individual $i$ at generation $G$, if the trial vector generated by its favor method $IP_{i,G}$ successfully replaces the target vector, then the preference is preserved to the next generation (line 7 in Algorithm 1). Otherwise, the individual $i$ will randomly reconnect to another distinct layer, in which another DE is employed (line 11 in Algorithm 1). Accordingly, algorithmic settings of the $M$ methods are updated, following their original designs (lines 8 and 12 in Algorithm 1). In this way, the $M$ layers compete and the winner will eventually take more individuals. An illustrative example is depicted in Fig. 1(b).

The parallel framework can effectively deal with self-adaptive [4, 22, 39] and adaptive [29, 33, 46] DEs even though they have a different structure. An example is given in Fig. 2, where an adaptive DE and a self-adaptive DE reside at $L_1$ and $L_2$, respectively (Note: Other layers are ignored for clarity.)

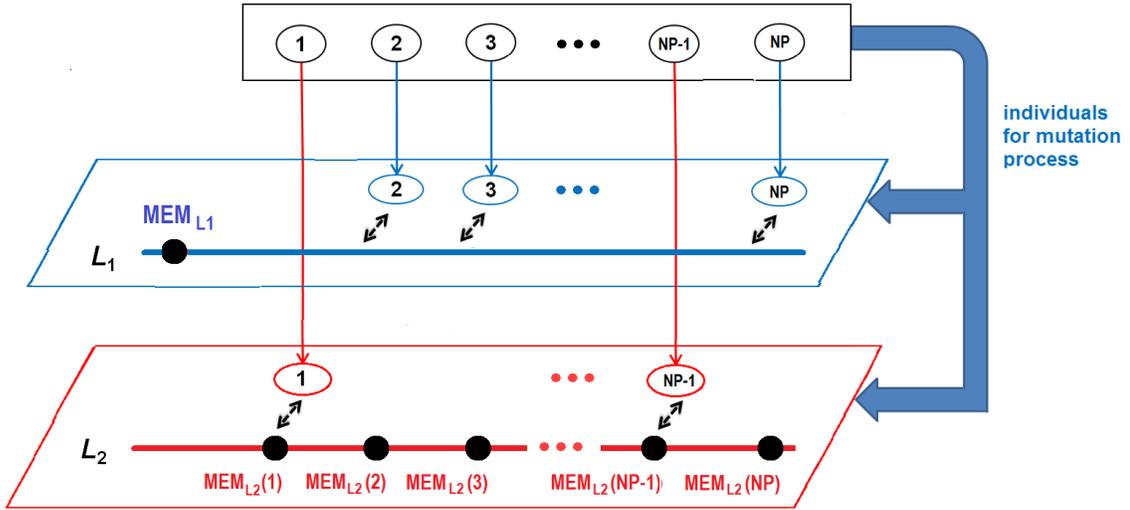

Fig.2 Illustration of parameter adaptation in multi-layer structure. For self-adaptive DE in $L_2$, strategy or (and) parameters corresponding to each individual are stored. For adaptive DE in $L_1$, only population-wise strategies or (and) parameters are needed. The double-headed arrows indicate the interaction between individuals and the memory in the layer.

For self-adaptive DEs, such as jDE [4], EPSDE [22] and CoBiDE [39], generational strategy or (and) parameters are associated with each individual. Therefore, as shown in the example, $L_2$ records and updates strategies or (and) parameters corresponding to each individual in memory $\{MEM_{L2}\{i\}, i \in \{1, 2, \cdots NP\}\}$ throughout the entire evolutionary process. When $\vec{x}_{i,G}$ is associated with $L_2$, i.e. $IP_{i,G} = 2$, strategies or (and) parameters stored in $MEM_{L2}\{i\}$ will be taken to generate a trial vector $\vec{u}_{i,G}$. Consequently, $MEM_{L2}\{i\}$ will be updated according to the fitness comparison result between $\vec{x}_{i,G}$ and $\vec{u}_{i,G}$.

For adaptive DEs, such as JADE [46] and SHADE [33], following their original designs, only population-wise strategies or (and) parameters are required. Given a layer using adaptive DEs, this piece of information will be stored and used by any individual associated with the layer. Consider the example shown in Fig. 2, SHADE is used in $L_1$. Hence, the memory $M_F$ and $M_{CR}$ [33] are used to store the values determined by the latest successful control parameters $F$ and $CR$. For any $\vec{x}_{i,G}$ with $IP_{i,G} = 1$, $L_1$ retrieves parameters from $M_F$ and $M_{CR}$ for the generation of trial vectors. Subsequently, parameters determined by the successful update of $\vec{x}_{i,G}$ are archived to update $M_F$ and $M_{CR}$.

In this paper, an example of combining two adaptive DEs is described in Sec. 4.1 while another example of integrating an adaptive DE and a self-adaptive DE is given in Sec. 4.4.

---------------------------------------------------------------------------------------------------------------------

**Algorithm 1.  Competitive IPLS mechanism**

---------------------------------------------------------------------------------------------------------------------

1: Set generation count $G = 0$, initialize a population $P_0 = \{\vec{x}_{i,0}, i \in \{1, 2, \cdots NP\}\}$, initialize each method $m$ ($m = 1$, 2, …, $M$), initialize individual preference $\{IP_{i,0} = ceil(rand_i(0,1) \times M), i \in \{1, 2, \cdots NP\}\}$. // $ceil(\cdot)$ *denotes a ceiling value.*

2: **While** the stopping criteria are not satisfied, **Do**

3: **For** $i = 1$: $NP$ **Do**

4: For $\vec{x}_{i,G}$, generate a trial vector $\vec{u}_{i,G}$ by its preference method $IP_{i,G}$;

5: **If** $f(\vec{u}_{i,G}) < f(\vec{x}_{i,G})$

6: $\quad \vec{x}_{i,G+1} = \vec{u}_{i,G}$;

7: $\quad IP_{i,G+1} = IP_{i,G}$;

8: Update the generation strategies and parameter settings of the method $IP_{i,G}$ if required by its original design;

9: **Else**

10: $\quad \vec{x}_{i,G+1} = \vec{x}_{i,G}$;

11: $\quad IP_{i,G+1} = ceil(rand_i(0,1) \times M) \setminus IP_{i,G}$;     // *"$\setminus$" implies exclusion of the previous method*

12: Update the generation strategies and parameter settings of the method $IP_{i,G}$ if required by its original design;

13: **End If**

14: **End For**

15: Evaluate the current evolution status and update the settings of the $M$ methods if required by their original designs;

16: $G = G + 1$;

17: **End While**

---------------------------------------------------------------------------------------------------------------------

### 3.3 Cooperative RAB mechanism

Existing DEs allocate equal amount of computational resources to each individual, regardless of its potential in finding a better solution. In the following experiment, it is shown that it is more likely to generate new best solutions (*NBS*) by evolving superior solutions than inferior ones. Therefore, an even distribution of resources may not be efficient.

The experiment is conducted with two classic algorithms "DE/rand/1/bin" and "DE/best/1/bin" tested with 30-dimensional CEC2014 benchmark functions. Parameters settings for both algorithms are: $F = 0.7$, $CR = 0.5$ and $NP = 5 \times D$, and the termination condition is set as $10^4 \times D$ function evaluations, where $D = 30$ is the dimension of the functions.

A rank archive, $R$, is used to record the rank of individuals who produce *NBS*, while $frequency_i$ indicates the frequency that the individual with $i$-th rank generates *NBS*. Define $AR$ as the average rank of individuals contributing to *NBS*, one has $AR = \frac{1}{r} \sum_{i=1}^{r} R_i$, where $r$ is the size of archive $R$. If the contribution of *NBS* is independent of individual's rank, the expected value of $AR$ can be computed by $AR^{exp} = (\sum_{i=1}^{NP} i) / NP = (\sum_{i=1}^{150} i) / 150 = 75.5$.

Fig. 3 depicts the value of $AR$ for each function in the median trial of 51 independent runs, while the dotted line indicates the value of $AR^{exp}$. As shown, $AR$ is smaller than $AR^{exp}$ on all the functions tested for both "DE/rand/1/bin" and "DE/best/1/bin". This simply implies that superior individuals have higher potentials to generate *NBS* than inferior ones. To further demonstrate this, Fig. 4 shows the value of $frequency_i$ on representative unimodal functions F2 and multimodal functions F9. It can be clearly seen that individuals with a higher rank produce more *NBS* than those with a lower rank.

Inspired by this phenomenon, a resource allocation bias (RAB) scheme is proposed to emphasize high-quality individuals by using all the layers. Its pseudocode is given in **Algorithm 2**. At each generation, the fitness ranking $FR(i)$ of each individual $i \in \{1, 2, \cdots NP\}$ is first determined (line 1). The smaller the $FR$, the better the solution. Then, the $top_G = ceil(rand(0,1) \times NP \times N)$ high-rank solutions are regarded as high-quality individuals, where $N \in [0,1]$ is a preset parameter. For each top rank individual, $M$ trial vectors $\vec{u}_{i^m,G}, m \in \{1, 2, \cdots M\}$ are generated by $M$ methods and the settings of each method are updated by comparing $\vec{u}_{i^m,G}$ with $\vec{x}_{i,G}$ (lines 4-8). Subsequently, the fittest trial vector $\vec{u}_{i,G}$ is chosen to compare with the target vector $\vec{x}_{i,G}$ (lines 9-14). In this way, the $M$ layers work cooperatively to promote the quality of the solution. On the contrary, inferior individuals are to produce one offspring each (line 16).

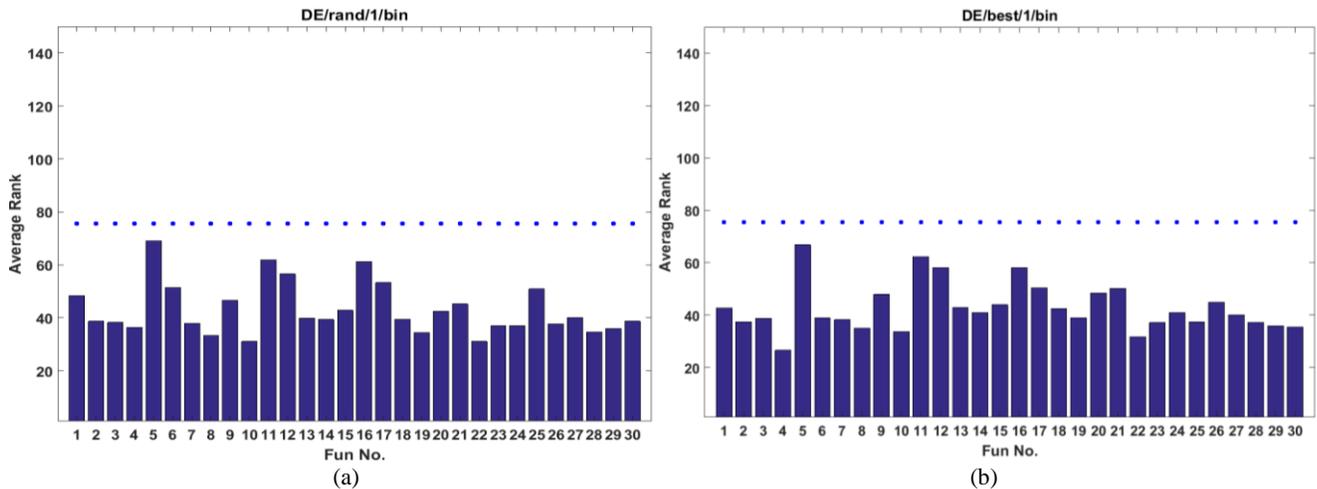

Fig. 3. Average Rank of individuals that generate new better solution (a) for "DE/rand/1/bin"; (b) for "DE/best/1/bin". Experiments are conducted on thirty 30-dimensional CEC2014 benchmark functions with 51 independent runs.

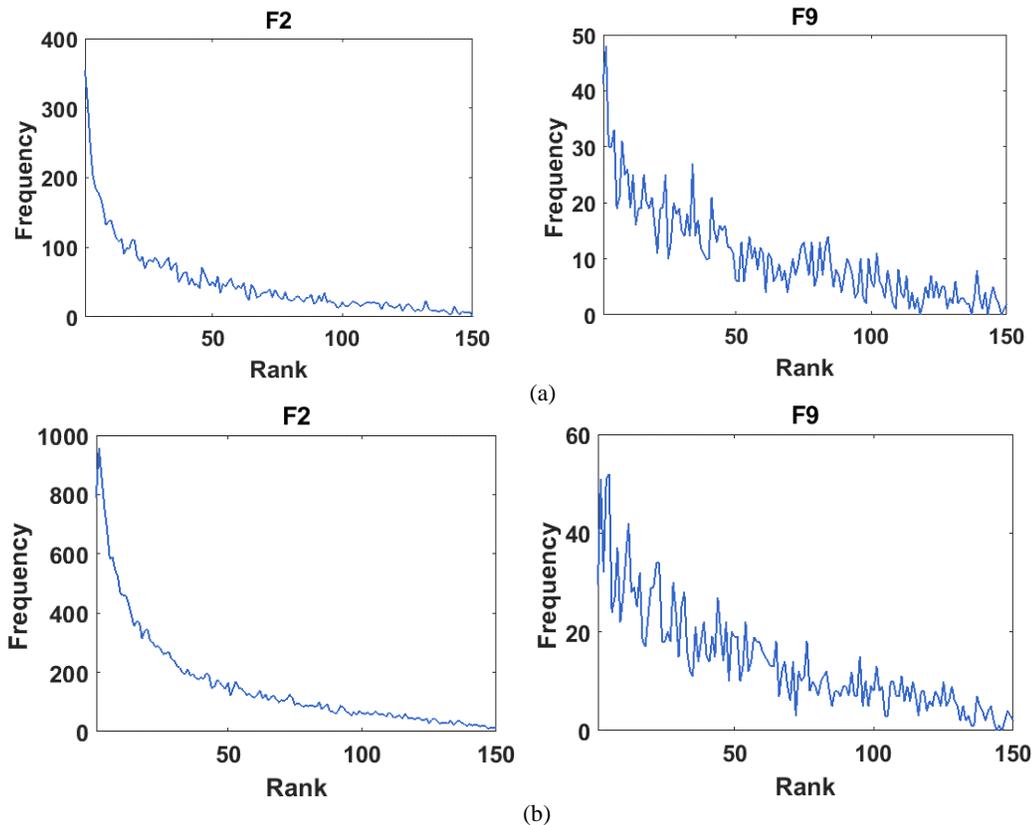

Fig. 4. Values of $frequency_i$ on unimodal functions F2 and multimodal functions F9: (a) for "DE/rand/1/bin"; (b) for "DE/best/1/bin".

The benefits of the cooperative RAB mechanism are twofold. Firstly, computational resources are re-distributed in a better way. At each generation, superior solutions are given $M$ trials by $M$ complementary methods. Therefore, the top individuals can be refined with a higher probability and are expected to lead the entire population towards more promising searching areas. Secondly, inferior solutions still have a chance to generate one candidate to compete with the superior solutions, thus maintaining the exploratory capability of DE.

---

**Algorithm 2. Cooperative RAB mechanism**

---

1: At generation $G$, determine the fitness ranking $FR(i)$, $i \in \{1, 2, \cdots NP\}$ of each individual, set $top_G = ceil(rand(0,1) \times NP \times N)$ ;

2: **For** $i = 1$: $NP$ **Do**

3:   **If** $FR(i) \leq top_G$

4:   For $\vec{x}_{i,G}$ , generate $M$ trial vectors $\vec{u}_{i^m,G}, m \in \{1, 2, \cdots M\}$ by using $M$ methods;

5:     **For** $m = 1$: $M$ **Do**

6:       Compare $\vec{u}_{i^m,G}$ with $\vec{x}_{i,G}$ ;

7:       Update the generation strategies and parameter settings of method $m$ if required by its original design;

8:     **End For**

9: Choose the best trial vector $\vec{u}_{i^b,G}$ in terms of fitness from $\vec{u}_{i^m,G}, m \in \{1, 2, \cdots M\}$ ,where $b$ indicates the index of the best method;

10:   **If** $f(\vec{u}_{i^b,G}) < f(\vec{x}_{i,G})$

11:     $\vec{x}_{i,G+1} = \vec{u}_{i^b,G}$ ;

12:   **Else**

13:     $\vec{x}_{i,G+1} = \vec{x}_{i,G}$ ;

14:   **End If**

15: **Else If** $FR(i) > top_G$

16:   For $\vec{x}_{i,G}$ , generate a trial vector $\vec{u}_{i,G}$ ;

17:   **If** $f(\vec{u}_{i,G}) < f(\vec{x}_{i,G})$

18:     $\vec{x}_{i,G+1} = \vec{u}_{i,G}$ ;

19:   **Else**

20:     $\vec{x}_{i,G+1} = \vec{x}_{i,G}$ ;

21:   **End If**

22: **End If**

23: **End For**

---

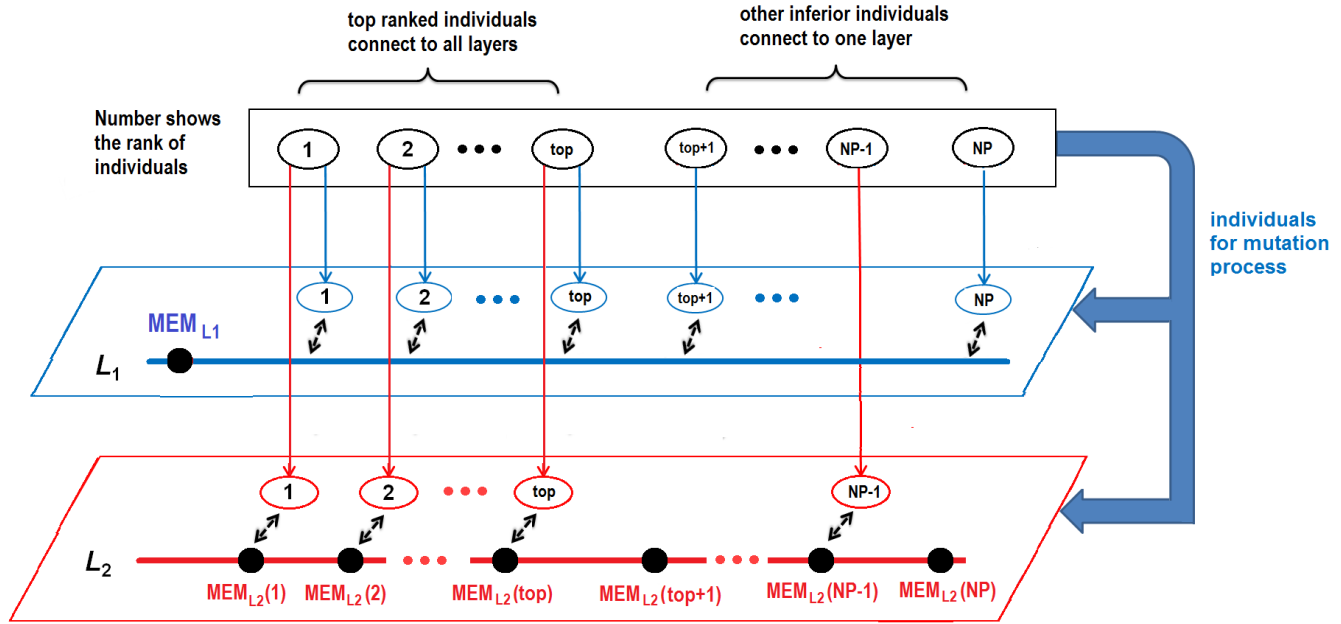

Fig.5 Illustration of MLCC ($M = 2$). The top ranked individuals connect to all layers and are evolved by M methods, while the others are connected to one layer. Note: For clarity, only two layers are shown and the number indicates the rank of an individual.

### 3.4 The MLCC Framework

Combining IPLS and RAB, the proposed MLCC framework is depicted in Fig. 5 and its pseudocode is presented in **Algorithm 3.** As observed, for target vectors with ranking $FR \leq top_G$, $M$ methods are considered (lines 6-9) and if the target vectors are successfully updated, their preferences are renewed with the corresponding best method $b$ (line 14). While for the inferior vectors, only their preferences are used (line 19).

Compared to existing methods, the novelty and characteristics of MLCC framework can be summarized by the following.

(1) The influence between individuals and each layer in MLCC is bidirectional. On the one hand, an individual can obtain algorithmic configurations from the layers for evolving, while on the other hand, it also returns feedbacks to the layers. This significantly differs from CoDE [38], in which algorithmic settings can only influence individuals regardless of the preference of each individual.

(2) MLCC introduces a novel multi-layer structure, which is in nature different from AMALGAM-SO [37], PAP [25], HDE [21] and MPEDE [44] which only uses one layer. With the multi-layer structure, each individual in MLCC can store, utilize and update its evolution information in multiple layers during the evolution, for example, they can evolve multiple layer-associated adaptive/self-adaptive $F$ and $CR$ parameters. Moreover, in MLCC, the incorporation of self-adaptive DEs [4, 22, 39] becomes much easier.

(3) The "multi-layer", rather than "multi-population" feature in MLCC significantly increase the flexibility in integrating DE variants with relatively complex proposals, such as the multi-population based IDE [36] algorithm.

(4) In MLCC, each layer has access to the current population. Although only part of the entire population is evolved by the $m$-th layer, where $m \in \{1, 2, \cdots M\}$, the vectors for mutation are selected from the entire population, following the original design of $m$-th method. In this context, the $M$ methods work in a collaborative manner. This is different from PAP [25] and MPEDE [44], in which individuals evolve only within their corresponding subpopulations.

(5) MLCC preserves the original design of the baselines. The procedures performed in each layer identical to those in the original algorithms, making MLCC easy to implement.

(6) RAB mechanism is introduced in MLCC to redistribute the computational resources and simultaneously take advantages of all the $M$ methods to enhance performance.

----------------------------------------------------------------------------------------------------

**Algorithm 3. The MLCC framework**

----------------------------------------------------------------------------------------------------

1: Initialize a population $P_0 = \{\vec{x}_{i,0}, i \in \{1, 2, \cdots NP\}\}$, initialize each method $m$ ($m$ = 1, 2, …, $M$), initialize the individual preference $\{IP_{i,0} = ceil(rand_i(0,1) \times M), i \in \{1, 2, \cdots NP\}\}$, set generation count $G$ =0, set threshold value $N$;

2: **While** the stopping criteria are not satisfied, **Do**

3: Determine the fitness ranking $FR(i)$ of each individual $i$ in the population, set $top_G = ceil(rand(0,1) \times NP \times N)$;

4: **For** $i$ = 1: $NP$ **Do**

5: **If** $FR(i) \leq top_G$

6: For $\vec{x}_{i,G}$, generate $M$ trial vectors $\vec{u}_{i^m,G}, m \in \{1, 2, \cdots M\}$ by using $M$ methods;

7: **For** $m$ = 1: $M$

8: Compare $\vec{u}_{i^m,G}$ with $\vec{x}_{i,G}$;

9: Update the generation strategies and parameter settings of method $m$ if required by its original design;

10: **End For**

11: Choose the best trial vector $\vec{u}_{i^b,G}$ in terms of fitness from $\vec{u}_{i^m,G}, m \in \{1, 2, \cdots M\}$, where $b$ indicates the index of the best method;

12: **If** $f(\vec{u}_{i^b,G}) \leq f(\vec{x}_{i,G})$

13: $\vec{x}_{i,G+1} = \vec{u}_{i^b,G}$;

14:    $IP_{i,G+1} = b$ ;

15:    **Else**

16:    $\vec{x}_{i,G+1} = \vec{x}_{i,G}$ ;

17:    **End If**

18:    **Else If** $FR(i) > top_G$

19:    For $\vec{x}_{i,G}$ , generate a trial vector $\vec{u}_{i,G}$ by its preference method $IP_{i,G}$ ;

20:    **If** $f(\vec{u}_{i,G}) \leq f(\vec{x}_{i,G})$

21:    $\vec{x}_{i,G+1} = \vec{u}_{i,G}$ ;

22:    $IP_{i,G+1} = IP_{i,G}$ ;

23:    Update the generation strategies and parameter settings of method $IP_{i,G}$ if required by its original design;

24:    **Else**

25:    $\vec{x}_{i,G+1} = \vec{x}_{i,G}$ ;

26:    $IP_{i,G+1} = ceil(rand_i(0,1) \times M) \setminus IP_{i,G}$ ;

27:    Update the generation strategies and parameter settings of the method $IP_{i,G}$ if required by its original design;

28:    **End If**

29:    **End If**

30:    **End For**

31:    Evaluate the current evolution status and update the settings of the $M$ methods if required by their original designs;

32:    $G = G + 1$;

33:    **End While**

-----------------------------------------------------------------------------------------------------------

## 3.5 On the Selection of the $M$ Methods

This subsection discusses the selection criteria of the $M$ methods for MLCC. In general, the following guidelines are given. 1) The $M$ methods are high-performers in order to construct a competitive DE; 2) the $M$ methods should complement each other to ensure a stable performance for a wide range of problems.

To determine suitable candidates, nine state-of-the-art and up-to-date DE variants, namely jDE [4], SaDE [29], EPSDE [22], JADE [46], CoDE [38], CoBiDE [39], MPEDE [44], SHADE [33] and IDE [36] have been run on 30-dimensional CEC2014 benchmark function set. The CEC2014 benchmark set is considered because it covers a wide range of functions with diverse mathematical properties. Therefore, the test results would reflect the overall performance of an algorithm.

Parameter settings for the DEs being considered, are summarized in Table S1 in the supplemental file. The mean and standard deviations of solution error values, given by $f(x) - f(x^*)$, over 51 independent runs are tabulated in Table S2 in the supplemental file, where $f(x^*)$ and $f(x)$ are the global optima and the best fitness after $10^4 \times D$ function evaluations, respectively [18]. The comparison results of the DEs given by Wilcoxon signed-rank test [32] with a significance level of 0.05 are summarized in Table S3.

In Table 1, the $p$-values obtained by comparing IDE with the other four most competitive DEs are presented, while the overall performance rankings of the nine considered DEs are summarized in Table 2.

As observed in Table 1, the performance of IDE is comparable to CoDE, CoBiDE, MPEDE and SHADE at $\alpha = 0.05$. As shown in Table 2, SHADE and IDE are the best and second best-performing DEs with ranking values of 3.48 and 3.53, respectively. In addition, according to single problem analysis between SHADE and IDE using Wilcoxon signed-rank test with 5% significance level, IDE wins, ties and loses in 13, 8 and 9 functions respectively when compared with SHADE. This indicates that the characteristics of SHADE and IDE complement each other. In summary, Tables 2 and S2 show that SHADE and IDE are the appropriate candidates for MLCC.

Table 1 P-values obtained by comparing IDE with the other four
most competitive DEs according to multi-problem Wilcoxon's test

| IDE v.s. | $R+$ | $R-$ | $p$-value | $\alpha = 0.05$ |
|---|---|---|---|---|
| CoDE | 263.5 | 171.5 | 0.314 | No |
| CoBiDE | 240.5 | 194.5 | 0.611 | No |
| MPEDE | 267.5 | 167.5 | 0.274 | No |
| SHADE | 271.0 | 194.0 | 0.422 | No |

Table 2 Performance ranking of the considered DE variants on 30-dimensional
CEC2014 benchmark set using Friedman's test

| Algorithm | Ranking |
|---|---|
| SHADE | **3.48** |
| IDE | 3.53 |
| CoBiDE | 4.06 |
| MPEDE | 4.21 |
| CoDE | 4.86 |
| JADE | 5.15 |
| jDE | 5.76 |
| EPSDE | 6.38 |
| SaDE | 7.53 |

### 3.6 The MLCC-SI Algorithm

Following **Algorithm 3**, the MLCC variant for two selected methods, SHADE and IDE, denoted as MLCC-SI is implemented and the pseudocode is provided as Algorithm S-1 in the supplemental file. It should be noted that procedures for SHADE and IDE used in the layers are identical to those in the original literature [33, 36].

## 4. Simulation and Discussion

In this section, the effectiveness of the proposed MLCC framework and the performance of the MLCC variants is verified through comprehensive experiments conducted on the CEC2014 test set [18]. The 30 benchmark functions in the CEC2014 test set can be classified into four categories: unimodal functions (F1-F3), simple multimodal functions (F4-F16), hybrid functions (F17-F22) and composition functions (F23-F30).

Performance of the considered algorithms is evaluated based on solution error value, which was defined previously in Sec. 3.5. Following the suggestion in [18], solution error values smaller than $10^{-8}$ are reported as zero. In the experiments, each algorithm is run independently on every function for 51 times. In each run, $10^4 \times D$ function evaluations are limited, while the final solution error values obtained are compared. It is noted that, to have a fair comparison, the initial populations for all algorithms are set to be the same as in a single run. In the tables presented, the best results achieved for each function is marked in **bold**.

To have statistically sound conclusions, single problem Wilcoxon's signed-rank test [32] with a significance level of 0.05, multiple problem Wilcoxon's test [11] and Friedman's test [11] are used in the performance comparison. Regarding single problem Wilcoxon's signed-rank test, the symbols "-", "=" and "+" in the tables represent that the performance of the compared algorithm is significantly worse than, similar to or better than that of the considered algorithm, respectively. In addition, for ease of comparison, "Positive subtracts Negative" value (P-N value) is also given, where "Positive" is the number of functions that the considered algorithm outperforms the algorithm compared while "Negative" is the number of functions for the opposite case.

### 4.1 Effectiveness of the MLCC Framework

In this subsection, the effectiveness of the proposed MLCC framework is verified through performance comparisons between the MLCC-SI algorithm and its two baseline DEs on the 30 and 50-dimensional CEC2014 test sets. Parameter settings for the algorithms are summarized as follows:

1) SHADE：$NP = 5 \times D$, $M_F = \{0.7\}$, $M_{CR} = \{0.5\}$, and $H = NP$.

2) IDE：$NP = 5 \times D$, $T = 1000D/NP$, $G_T = 5T$, $SR_T = 0$ $(G < G_T)$, and $SR_T = 0.1$ $(G \geq G_T)$ .

3) MLCC-SI：$M_F = \{0.7\}$, $M_{CR} = \{0.5\}$, and $H = NP$ (For SHADE layer); $T = 1000D/NP$, $G_T = 5T$, $SR_T = 0$ $(G < G_T)$, and $SR_T = 0.1$ $(G \geq G_T)$ (For IDE layer); $NP = 5 \times D$ and $N = 0.05$.

The mean and standard deviations of error values achieved with 51 independent runs and the statistical comparison results are shown in Table 3.

From Table 3, it can be observed that MLCC-SI performs significantly better than SHADE and IDE. Out of the total 120 cases, MLCC-SI wins in 74 (=15+16+22+21) cases and only loses in 12 (=5+1+4+2) cases. MLCC-SI outperforms SHADE in 37 (=15+22) functions and underperforms in 9 (=5+4) functions. When compared with IDE, MLCC-SI is superior in 37 (=16+21) cases and inferior in 3 (=1+2) cases.

Considering the features of the test functions, the following results can be observed:

For unimodal functions F1-F3, SHADE performs the best while IDE is the worst. MLCC-SI loses to SHADE in 3 cases but wins IDE in 4 cases.

For simple multimodal functions F4-F16, MLCC-SI significantly outperforms SHADE and IDE. In the total 52 (=13$\times$4) cases, MLCC-SI wins SHADE and IDE in 18 (=8+10) and 15 (=7+8) cases and loses in 2 (=2+0) and 1 case, respectively.

For hybrid functions F17-F22, Table 3 shows that MLCC-SI is again the best. MLCC-SI performs better than SHADE and IDE in 21 functions and only loses in 1 function.

For composition functions F23-F30 with complex mathematical characteristics, from Table 3, MLCC-SI is also the best performer. It is superior to SHADE and IDE in 9 (3+6) and 7 (=2+5) cases and inferior in 3 (=1+2) and 2 (=1+1) cases, respectively.

Furthermore, the performance of MLCC-SI, SHADE, and IDE are compared according to multiple problem Wilcoxon's test, and the results are shown in Table 4. Regarding the *p*-value obtained, it can be concluded that the overall performance of MLCC-SI is significantly better than those of SHADE and IDE with 5% significance level. This is also confirmed by the Friedman's test results, as given in Table 5, that MLCC-SI achieves a much smaller ranking value (1.45) while SHADE and IDE perform similarly. In conclusion, MLCC significantly improves the performance of the baseline DEs.

## 4.2 Benefits of the Components in MLCC

This subsection studies the advantages of the two components i.e. IPLS and RAB mechanisms designed in MLCC. Four variants, denoted as Variants I-IV of MLCC-SI are constructed as follows.

*Variant-I:* MLCC-SI without RAB. In this variant, at each generation, each individual can connect to only one layer based on its preference.

*Variant-II*: MLCC-SI without IPLS. In this variant, at each generation, the superior $top_G$ individuals connect to $M$ layers while the remains randomly connect to only one layer.

*Variant-III*: MLCC-SI without IPLS and RAB. In this variant, at each generation, each individual randomly connects to only one layer.

*Variant-IV*: MLCC-SI without fitness bias. In this variant, the $top_G$ individuals permitted to connect to $M$ layers are randomly selected from the entire population without fitness bias.

Table 3 Performance comparisons of MLCC-SI with its baseline DE variants on 30- and 50-dimensional cec2014 benchmark set over 51 independent runs

| | *D* = 30 | | | *D* = 50 | | |
|---|---|---|---|---|---|---|
| | SHADE | IDE | MLCC-SI | SHADE | IDE | MLCC-SI |
| F1 | **2.59E+02** + (5.67E+02) | 1.18E+05 - (9.41E+04) | 4.76E+03 (5.69E+03) | **1.19E+05** - (6.14E+04) | 1.24E+06 - (3.41E+05) | 2.79E+05 (1.00E+05) |
| F2 | **0.00E+00** = (0.00E+00) | **0.00E+00** = (0.00E+00) | **0.00E+00** (0.00E+00) | **0.00E+00** + (0.00E+00) | 2.28E+00 - (2.53E+00) | 2.67E-04 (3.59E-04) |
| F3 | **0.00E+00** = (0.00E+00) | **0.00E+00** = (0.00E+00) | **0.00E+00** (0.00E+00) | **0.00E+00** = (0.00E+00) | 1.85E+01 - (1.27E+01) | 2.10E-10 (1.50E-09) |
| F4 | **0.00E+00** + (0.00E+00) | 2.08E-02 - (4.14E-02) | 1.63E-07 (4.37E-07) | 8.35E+01 - (1.16E+01) | 7.19E+01 - (2.97E+01) | **6.53E+01** (2.62E+01) |
| F5 | 2.03E+01 - (3.54E-02) | **2.02E+01** - (5.68E-02) | **2.02E+01** (5.40E-02) | 2.05E+01 - (4.03E-02) | **2.03E+01** = (5.95E-02) | **2.03E+01** (5.46E-02) |
| F6 | 6.41E+00 - (3.86E+00) | **6.20E-02** = (2.82E-01) | 8.71E-02 (2.84E-01) | 1.18E+00 = (3.45E+00) | **9.34E-02** + (3.14E-01) | 3.96E-01 (5.61E-01) |
| F7 | **0.00E+00** = (0.00E+00) | **0.00E+00** = (0.00E+00) | **0.00E+00** (0.00E+00) | **0.00E+00** = (0.00E+00) | 2.22E-03 - (4.10E-03) | **0.00E+00** (0.00E+00) |
| F8 | **0.00E+00** = (0.00E+00) | 4.33E-10 - (3.09E-09) | **0.00E+00** (0.00E+00) | 1.84E-02 - (5.39E-03) | 4.32E-02 - (1.97E-01) | **0.00E+00** (0.00E+00) |
| F9 | 2.75E+01 - (4.18E+00) | 2.46E+01 - (5.33E+00) | **2.14E+01** (4.44E+00) | 8.82E+01 - (8.25E+00) | 5.99E+01 - (1.01E+01) | **4.47E+01** (8.15E+00) |
| F10 | **1.57E-01** - (3.94E-02) | 5.68E+00 - (1.66E+01) | 1.12E+00 (9.49E-01) | 6.06E+01 - (6.43E+00) | 3.34E+01 - (4.90E+01) | **9.00E+00** (3.38E+00) |
| F11 | 1.97E+03 - (2.06E+02) | 1.92E+03 - (3.53E+02) | **1.63E+03** (3.34E+02) | 6.27E+03 - (3.93E+02) | 4.20E+03 - (6.65E+02) | **4.03E+03** (5.06E+02) |
| F12 | 3.08E-01 - (4.82E-02) | 2.91E-01 - (5.97E-02) | **2.60E-01** (5.31E-02) | 6.12E-01 - (6.73E-02) | 3.68E-01 - (7.37E-02) | **3.51E-01** (5.92E-02) |
| F13 | 2.15E-01 - (2.58E-02) | 1.87E-01 - (2.20E-02) | **1.83E-01** (2.79E-02) | 3.01E-01 - (2.99E-02) | 2.96E-01 - (3.09E-02) | **2.77E-01** (2.58E-02) |
| F14 | 2.14E-01 - (2.24E-02) | **1.82E-01** - (3.19E-02) | 1.94E-01 (2.21E-02) | **2.50E-01** = (1.82E-02) | 2.70E-01 - (2.23E-02) | 2.56E-01 (2.36E-02) |
| F15 | 3.83E+00 - (4.70E-01) | 2.69E+00 - (5.27E-01) | **2.47E+00** (4.20E-01) | 1.18E+01 - (8.02E-01) | 7.36E+00 - (1.93E+00) | **6.41E+00** (1.34E+00) |
| F16 | 9.55E+00 - (3.49E-01) | 1.00E+01 - (3.94E-01) | **9.52E+00** (4.66E-01) | 1.88E+01 - (2.77E-01) | 1.92E+01 - (4.21E-01) | **1.85E+01** (4.53E-01) |
| F17 | 7.62E+02 - (3.58E+02) | 5.97E+02 - (2.97E+02) | **2.31E+02** (1.23E+02) | 2.21E+03 - (5.57E+02) | 7.22E+03 - (2.74E+03) | **1.27E+03** (4.01E+02) |
| F18 | 1.44E+01 - (7.28E+00) | 1.90E+01 - (5.87E+00) | **9.79E+00** (3.36E+00) | 8.03E+01 - (2.31E+01) | 3.93E+01 - (1.09E+01) | **3.55E+01** (1.17E+01) |
| F19 | 4.01E+00 - (6.47E-01) | **2.91E+00** - (4.69E-01) | 3.02E+00 (5.37E-01) | 1.29E+01 - (5.85E+00) | 1.03E+01 - (7.50E-01) | **9.87E+00** (3.98E-01) |
| F20 | **4.96E+00** + (2.19E+00) | 1.08E+01 - (3.24E+00) | 5.91E+00 (1.42E+00) | 4.11E+01 - (1.63E+01) | 4.54E+01 - (1.04E+01) | **2.53E+01** (6.78E+00) |
| F21 | 1.29E+02 = (8.62E+01) | 3.30E+02 - (1.54E+02) | **1.04E+02** (7.65E+01) | 9.75E+02 - (2.81E+02) | 1.23E+03 - (3.77E+02) | **5.42E+02** (1.92E+02) |
| F22 | 1.23E+02 - (5.85E+01) | 7.30E+01 - (5.78E+01) | **3.55E+01** (3.45E+01) | 4.85E+02 - (1.22E+02) | 3.04E+02 = (1.06E+02) | **2.75E+02** (1.13E+02) |
| F23 | **3.15E+02** = (4.02E-13) | **3.15E+02** + (3.46E-13) | **3.15E+02** (4.02E-13) | **3.44E+02** = (4.60E-13) | **3.44E+02** = (4.46E-13) | **3.44E+02** (4.18E-13) |
| F24 | **2.23E+02** = (9.22E-01) | **2.23E+02** = (7.24E-01) | **2.23E+02** (7.91E-01) | 2.69E+02 - (1.90E+00) | **2.58E+02** = (3.39E+00) | **2.58E+02** (2.93E+00) |
| F25 | 2.04E+02 - (7.68E-01) | **2.03E+02** = (2.33E-01) | **2.03E+02** (2.95E-01) | 2.11E+02 - (2.59E+00) | 2.07E+02 - (6.05E-01) | **2.06E+02** (8.22E-01) |
| F26 | **1.00E+02** = (2.79E-02) | **1.00E+02** = (2.60E-02) | **1.00E+02** (2.41E-02) | **1.00E+02** = (3.37E-02) | 1.06E+02 = (2.37E+01) | **1.00E+02** (2.83E-02) |
| F27 | **3.00E+02** + (1.11E-13) | 3.30E+02 - (4.63E+01) | 3.47E+02 (5.07E+01) | 3.33E+02 - (2.79E+01) | **3.06E+02** = (1.65E+01) | 3.20E+02 (2.65E+01) |
| F28 | 7.92E+02 - (1.86E+01) | 8.26E+02 - (8.10E+01) | **7.89E+02** (3.09E+01) | **1.09E+03** + (3.20E+01) | 1.28E+03 - (9.49E+01) | 1.16E+03 (3.60E+01) |
| F29 | 7.20E+02 - (6.01E+00) | **5.75E+02** - (2.15E+02) | 6.94E+02 (1.27E+02) | 8.27E+02 - (5.63E+01) | 1.03E+03 - (1.26E+02) | **6.22E+02** (1.41E+02) |
| F30 | 1.22E+03 - (4.61E+02) | **5.18E+02** - (7.28E+01) | 5.20E+02 (1.60E+02) | **8.45E+03** + (4.59E+02) | 9.90E+03 - (5.82E+02) | 8.61E+03 (3.99E+02) |
| -/=/+ | **15/10/5** | **16/13/1** | | **22/4/4** | **21/7/2** | |

Table 4 Comparison results of MLCC-SI with its baseline DE variants on 30- and 50-dimensional benchmark set according to multi-problem Wilcoxon's test

| MLCC-SI **v.s.** | $R+$ | $R-$ | $p$-value | $\alpha = 0.05$ |
|---|---|---|---|---|
| SHADE | 1394.5 | 375.5 | **1.18E-04** | **Yes** |
| IDE | 1512.0 | 258.0 | **2.00E-06** | **Yes** |

Table 5 Overall performance ranking of MLCC-SI and its baseline DE variants on 30- and 50-dimensional CEC2014 benchmark set by Friedman's test

| Algorithm | Ranking |
|---|---|
| MLCC-SI | **1.45** |
| IDE | 2.26 |
| SHADE | 2.28 |

Parameter settings for these variants are set the same as those for MLCC-SI, as summarized in Sec. 4.1. Their performance comparisons with MLCC-SI are presented in Table S4 in the supplemental file and summarized in Table 6. As shown in Table 6, MLCC-SI performs better than all the variants in both 30 and 50-dimensional cases. The effectiveness of RAB, IPLS, the overall performance contributions by RAB and IPLS, and the benefit of fitness bias can be observed by comparing MLCC-SI with Variants I-IV, respectively. To show further the performance improvements on the baseline DEs, the performance of the four variants are also compared with SHADE and IDE, as shown in Tables S5-S8 in the supplemental file and Table 7. It can be seen that, with respect to the "-/=/+" results and P-N values, MLCC-SI exhibits higher improvements than the four variants. Considering the total P-N values achieved by the five algorithms, Table 7 shows that MLCC-SI performs the best with the maximum P-N value (62) while Variant-III without IPLS and RAB is the worst.

It is interesting to investigate the individual preferences at different evolution stages. To this end, the entire searching process is divided into several non-overlapping intervals, each consists of 50 generations. Figure 6 plots the evolution of the preference of four randomly selected initial individuals to the two layers SHADE and IDE in three typical 50-dimensional CEC2014 benchmark functions, namely F13 (multimodal function), F17 (hybrid function) and F23 (composition function). It is observed that 1) for function F13, Fig. 6(a) indicates that the four individuals have quite different preferences. The relative percentage $P_i$ processed by different layers of these four individuals varies at the same searching stages; 2) for function F17, the individuals demonstrate similar preferences throughout the entire process (Fig. 6(b)); 3) for function F23, Fig. 6(c) shows that all the individuals favor more to the IDE layer at the early stage, but vary at the later stage.

Fig. 7(a) compares the average rank $AR$ (defined in Sec. 3.3) of MLCC-SI with that of Variant-I, while Fig.7(b) shows the $AR$ of MLCC-SI, SHADE and IDE, for the thirty 50-dimensional CEC2014 functions. It is observed from Fig. 7(a) that MLCC-SI achieves smaller $AR$ values than Variant-I on all the functions, which means that the RAB mechanism enables MLCC-SI to focus more on superior individuals. Similarly, Fig. 7(b) shows that MLCC-SI also achieves smaller $AR$ values than SHADE and IDE on all the functions except functions F3 and F8, which indicates that MLCC-SI emphasizes more on high-quality solutions compared with SHADE and IDE.

Table 6 Comparison results of MLCC-SI with its four variants on 30 and 50-dimensional CEC2014 benchmark set according to Wilcoxon signed-rank test with a significance level of 0.05

| -/=/+ | $D = 30$ | $D = 50$ |
|---|---|---|
| Variant-I | 8/21/1 | 7/21/2 |
| Variant-II | 8/19/3 | 13/15/2 |
| Variant-III | 13/14/3 | 16/12/2 |
| Variant-IV | 7/21/2 | 7/19/4 |

Table 7 Comparison results of MLCC-SI and its four variants with the baseline des on 30- and 50-dimensional cec2014 benchmark set according to Wilcoxon signed-rank test with a significance level of 0.05

| -/=/+ (P-N) | v.s. | $D = 30$ | $D = 50$ | Total P-N value |
|---|---|---|---|---|
| Variant-I | SHADE | 12/11/7 (5) | 18/5/7 (11) | (37) |
| | IDE | 13/13/4 (9) | 18/6/6 (12) | |
| Variant-II | SHADE | 15/11/4 (11) | 18/8/4 (14) | (42) |
| | IDE | 13/13/4 (9) | 15/8/7 (8) | |
| Variant-III | SHADE | 13/11/6 (7) | 16/9/5 (11) | (25) |
| | IDE | 9/15/6 (3) | 15/4/11 (4) | |
| Variant-IV | SHADE | 14/9/7 (7) | 18/6/6 (12) | (41) |
| | IDE | 14/11/5 (9) | 18/7/5 (13) | |
| MLCC-SI | SHADE | 15/10/5(10) | 22/4/4 (18) | **(62)** |
| | IDE | 16/13/1 (15) | 21/7/2 (19) | |

Table 8 Comparison results of different settings on 30-dimensional cec2014 benchmark set according to Wilcoxon signed-rank test with a significance level of 0.05

| | -/=/+ | | -/=/+ |
|---|---|---|---|
| Setting-I | 2/27/1 | Setting-IV | 9/18/3 |
| Setting-II | 2/26/2 | Setting-V | 4/25/1 |
| Setting-III | 6/21/3 | Setting-VI | 11/15/4 |

## 4.3 Performance Sensitivity to $N$

This subsection investigates the performance sensitivity of MLCC-SI to its parameter $N$ by comparing the standard MLCC-SI with $N = 0.05$ with four other settings, i.e. Settings I-IV with $N = 0.1, 0.2, 0.5$ and $1.0$, respectively. Besides, two more settings, i.e. Settings V and VI with extreme settings of $top_G = 1$ and $top_G = NP$, respectively, are also considered. Performance comparisons on 30-dimensional CEC2014 functions are tabulated in Table S9 and summarized in Table 8.

According to Table 8, the followings can be concluded: 1) the performance of Settings-I and II are comparable to that of MLCC-SI, implying that MLCC-SI is robust when $N$ is small, such as 0.05, 0.1 or 0.2; 2) the performance of Settings III and IV is inferior to that of MLCC-SI, indicating that $N$ values that are too large will deteriorate the performance; 3) when comparing the performance of Settings-V and VI with that of MLCC-SI, the cases for $top_G = 1$ and $top_G = NP$ did not perform as well as MLCC-SI. In general, a larger $top_G$ value enables more superior solutions to be improved. However, when $top_G$ is too large, e.g. $top_G = NP$ in Setting-VI, the computation resources are again uniformly distributed and the performance benefit less from evolving the inferior solutions.

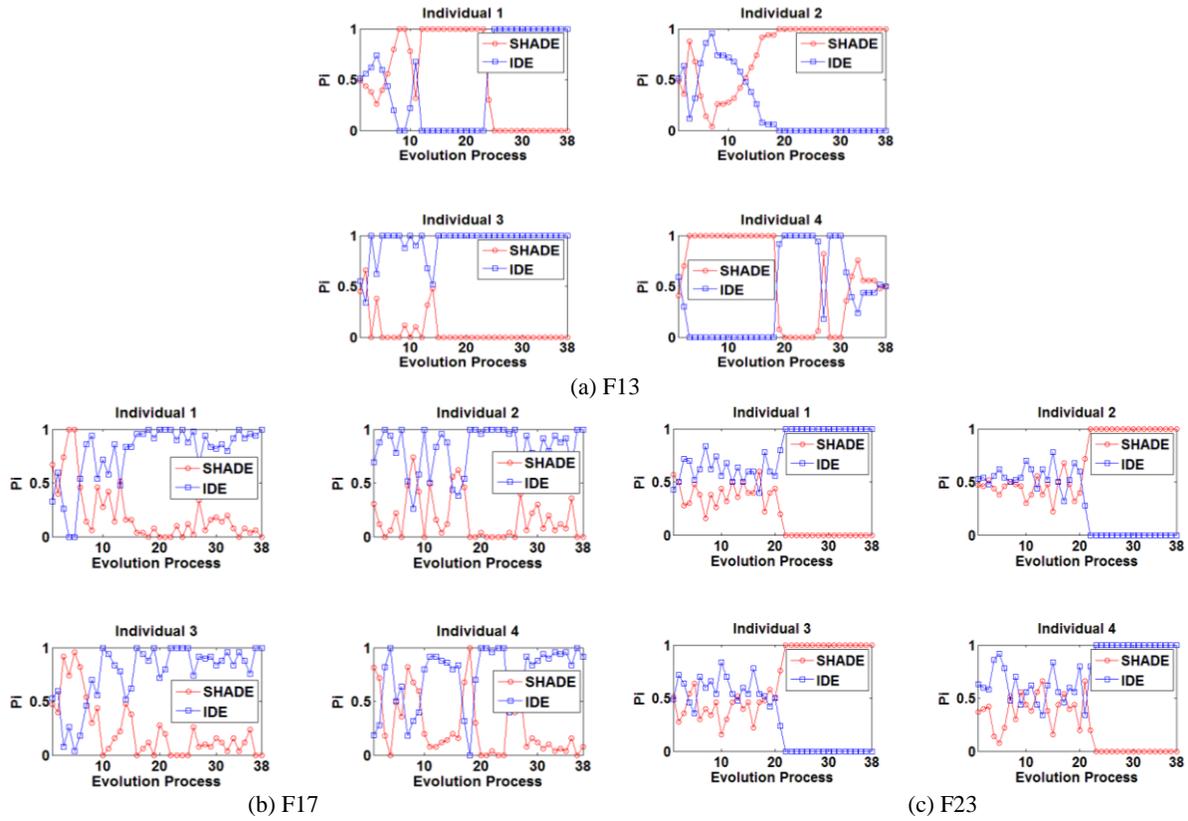

(a) F13

(b) F17                                                              (c) F23

Fig. 6 Evolution of the relative percentage *Pi* processed by different layers of four randomly selected initial individuals on three 50-dimensional CEC2014 benchmark functions F13, F17 and F23 in the median run.

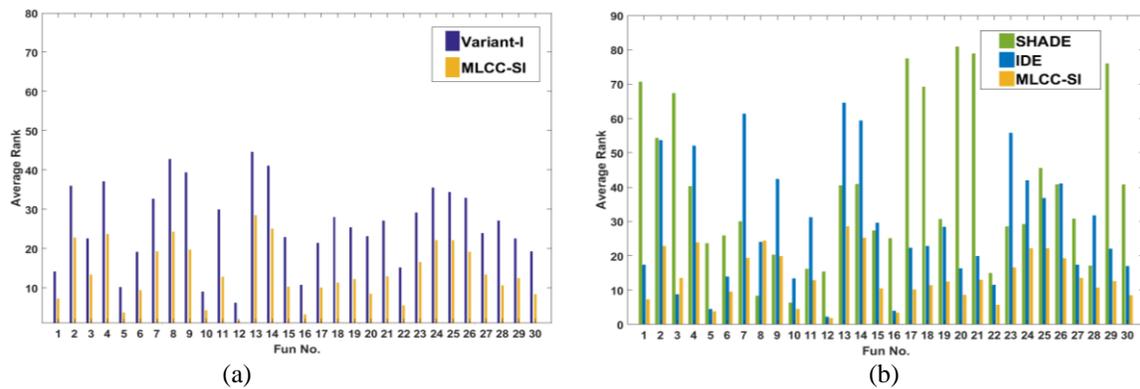

(a)                                                                    (b)

Fig.7 Average rank on thirty 50-dimensional CEC2014 benchmark functions: (a) MLCC-SI and Variant-I; (b) MLCC-SI, SHADE and IDE.

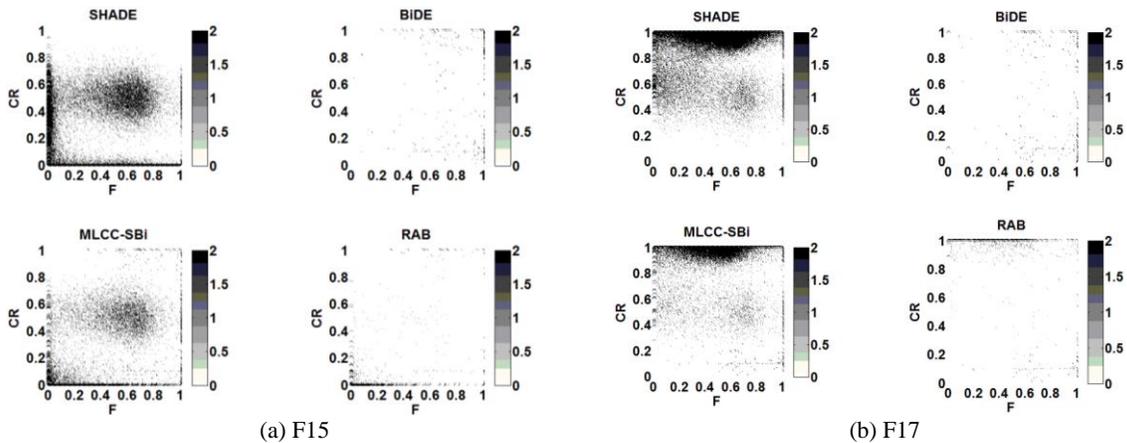

(a) F15                                                                                    (b) F17

Fig. 8 Distribution of the successful parameters *F* and *CR* generated by SHADE, BiDE, MLCC-SBi and RAB mechanism in MLCC-SBi on two 50-dimensional CEC2014 functions F15 and F17 in the median run. The darker, the higher frequency.

### 4.4 MLCC for Multi-Parameter Strategy Adaptation

Very recently, Tanabe and Fukunaga [35] investigated the behavior and performance of different parameter adaptation strategies [4, 22, 33, 39, 46] proposed in DE literature. They concluded, [35] by pointing out that "*there is still significant room for improvement in parameter adaptation methods for DE*".

Here, this subsection demonstrates the possibility of improving the performance of DE by the cooperation of multiple parameter adaptation strategies under the proposed MLCC framework. The adaptive Success History-based Parameter Configuration (SHA) originated from SHADE [33] and Self-adaptive Bimodal Distribution Parameter Scheme (BiD) derived from CoBiDE [39] are considered due to their competitive performance and representative characteristics. Two baseline DEs, assigned to two layers are designed as follows.

SHADE: the original SHADE algorithm [33];

BiDE: SHADE with SHA replaced by BiD [39].

In this way, by comparing the performance of the MLCC variant, i.e. MLCC-SBi, with that of SHADE and BiDE, the effectiveness of MLCC in multiple strategies adaptation can be observed. The pseudocode of MLCC-SBi is presented in Algorithm S-2 in the supplemental file. Parameter settings for the algorithms are summarized as follows.

1) SHADE：$NP = 5 \times D$, $M_F = \{0.7\}$, $M_{CR} = \{0.5\}$, and $H = NP$.

2) BiDE：$NP = 5 \times D$.

3) MLCC-SBi：$NP = 5 \times D$, $M_F = \{0.7\}$, $M_{CR} = \{0.5\}$, $H = NP$, and $N = 0.05$.

The experimental results on 30 and 50-dimensional CEC2014 test suite are presented in Table S10 and a summary is given in Table 9. As shown in Table S10 and Table 9, MLCC-SBi performs significantly better than SHADE and BiDE in both 30 and 50-dimensional cases. Specifically, MLCC-SBi wins the baseline DEs in 55 (=15+15+6+19) cases and loses in 10 cases (=2+2+3+3). Moreover, Table 9 shows that the superiority of MLCC-SBi over BiDE is more significant in the 50-dimensional case than in the 30-dimensional case. The reason is that SHA is not comparable to BiD. As shown in Table S11, BiDE outperforms SHADE with the "-/=/+" result of "14/10/6" in the 30-dimensional case. However, when the problem dimension increases to 50, the performance of BiD becomes comparable to that of SHA with "-/=/+" of "10/7/13", as indicated in Table S11. It should also be stressed that MLCC-SBi consistently exhibits better performance than both of the baseline algorithms. The cases that MLCC-SBi loses to BiDE are functions F24, F25, and F29 in both 30 and 50 dimensions. On the other functions, MLCC-SBi demonstrates significantly better or similar performance compared to BiDE.

To investigate factors that contribute to the performance improvements, the distribution of successful parameters $F$ and $CR$ associating with successful updates of the target vectors generated by SHADE, BiDE, MLCC-SBi and RAB mechanism in MLCC-SBi on two 50-dimensional CEC2014 functions F15 and F17 are plotted in Fig. 8. It can be seen that MLCC-SBi produces more diverse successful parameters than single SHA and BiD. The successful parameters generated by RAB come from both SHA and BiD, revealing that the proposed RAB mechanism simultaneously takes advantages of both schemes. To conclude, MLCC provides an effective approach to integrate multiple parameter adaptation schemes.

Table 9 Comparison results of MLCC-SBi with its baseline DE variants on 30- and 50-dimensional CEC2014 benchmark set according to Wilcoxon signed-rank test with a significance level of 0.05

| -/=/+ | $D = 30$ | $D = 50$ |
|---|---|---|
| SHADE | 15/13/2 | 15/13/2 |
| BiDE | 6/21/3 | 19/8/3 |

## 4.5 MLCC Versus Other Framework

To further demonstrate the superiority of the proposed MLCC framework, another very recently proposed hybrid DE framework, called HDE [21] is compared. In HDE, two algorithms are performed alternatively according to their fitness improvement rate. At each generation, only one algorithm is executed. When it did not perform well for several generations, another would be used. In this subsection, HDE is applied to SHADE and IDE, SHADE and BiDE, respectively, denoted as H-SI and H-SBi. Their performance are compared with those of MLCC-SI and MLCC-SBi, respectively. Parameter settings for the HDE framework are set the same as recommended in the original literature, while parameter settings for the baseline DEs and the MLCC framework are the same as those used previously in Sec. 4.1 and Sec. 4.4.

As seen from Table S12 and Table 10, MLCC framework exhibits better performance than HDE framework on both 30 and 50-dimensional functions. In the total of 120 cases, MLCC wins in 46 (=10+14+8+14) cases and loses in 14 (=6+4+3+1) cases. There may be two reasons that MLCC outperforms HDE. On the one hand, MLCC has the entire population monitored by multiple layers, which are performed simultaneously at each generation. Thus, individuals in MLCC could quickly respond to the change of evolution stage. While on the other hand, the RAB mechanism proposed in MLCC simultaneously takes advantages of multiple layers and also re-distributes the computation resources to help the algorithm focus more on promising searching directions.

Table 10 Comparison results of MLCC framework with
HDE framework on 30- and 50-dimensional CEC2014
benchmark set according to Wilcoxon signed-rank test with
a significance level of 0.05

| -/=/+ | $D = 30$ | $D = 50$ |
|---|---|---|
| MLCC-SI v.s. H-SI | 10/14/6 | 14/12/4 |
| MLCC-SBi v.s. H-SBi | 8/19/3 | 14/15/1 |

## 4.6 Comparisons with State-of-the-Art and Up-to-Date DEs

The effectiveness of the proposed MLCC framework have been verified in previous subsections. In this subsection, the MLCCDE algorithm based on SHADE and IDE and the following parameter settings, is compared with eight well-known state-of-the-art and up-to-date DEs, namely, jDE [4], SaDE [29], EPSDE [22], JADE [46], CoDE [38], CoBiDE [39], SinDE [9] and MPEDE [44].

Parameter settings of MLCCDE: $NP$ =100 (for $D = 30$) , $NP$ =150 (for $D = 50$) , $M_F = \{0.7\}$, $M_{CR} = \{0.5\}$, $H = NP$, $T = 1000D/NP$, $G_T = 5T$, $SR_T = 0$ ($G < G_T$), $SR_T = 0.1$ ($G \geq G_T$), and $N = 0.05$.

It is noted that MLCCDE uses different $NP$ settings from those of MLCC-SI. This is because MLCCDE empirically exhibits better overall performance with these settings, as compared to other DE variants. Parameter settings for the compared DEs are set the same as those given in their original literature. Here, the experiment also includes the recent CEC2017 test suite [3], in which several new functions are introduced.

The performance comparisons on 30 and 50-dimensional CEC2014 and CEC2017 functions are reported in Tables S13 -S16, and the comparison results are summarized in Table 11.

From Table 11, it can be observed that MLCCDE performs much better than the compared DEs. More specifically, in the CEC2014 30-dimensional case, MLCCDE outperforms jDE, SaDE, EPSDE, JADE, CoDE, CoBiDE, SinDE and MPEDE in 20, 27, 20, 18, 15, 13, 16 and 16 functions and underperforms in 2, 0, 5, 2, 4, 3, 5 and 6 functions, respectively. In the CEC2014 50-dimensional case, MLCCDE wins jDE, SaDE, EPSDE, JADE, CoDE, CoBiDE, SinDE and MPEDE in 22, 29, 23, 21, 22, 22, 18 and 17 functions respectively and loses in far fewer functions. For the CEC2017 functions, MLCCDE also exhibits much better performance than the compared DEs, as confirmed by the results in Table 11.

Considering multiple problems Wilcoxon's test, Tables 12 and 13 show that MLCCDE consistently achieves much larger $R+$ than $R-$ when compared with other DEs. The $p$-values obtained also confirm that MLCCDE significantly outperforms all the compared DEs at $\alpha = 0.05$. In addition, from the Friedman's test results shown in Table 14, MLCCDE achieves the smallest ranking values of 2.78 and 2.49 on CEC2014 and CEC2017 functions, respectively.

Table 11 Comparison results of MLCCDE with start-of-the-art and up-to-date DE variants on 30- and 50-dimensional CEC2014 and CEC2017 benchmark set according to Wilcoxon signed-rank test with a significance level of 0.05

| -/=/+ | CEC2014 | | CEC2017 | |
|---|---|---|---|---|
| | $D = 30$ | $D = 50$ | $D = 30$ | $D = 50$ |
| jDE | 20/8/2 | 22/6/2 | 21/9/0 | 23/5/2 |
| SaDE | 27/3/0 | 29/1/0 | 26/4/0 | 27/3/0 |
| EPSDE | 20/5/5 | 23/1/6 | 23/4/3 | 22/3/5 |
| JADE | 18/10/2 | 21/4/5 | 22/6/2 | 21/4/5 |
| CoDE | 15/11/4 | 22/4/4 | 15/13/2 | 23/6/1 |
| CoBiDE | 13/14/3 | 22/4/4 | 15/10/5 | 24/4/2 |
| SinDE | 16/9/5 | 18/7/5 | 18/10/2 | 19/5/6 |
| MPEDE | 16/8/6 | 17/9/4 | 15/8/7 | 17/10/3 |

Table 12 Comparison results of MLCCDE with start-of-the-art and up-to-date DE variants on 30- and 50-dimensional CEC2014 benchmark set according to multi-problem Wilcoxon's test

| MLCCDE **v.s.** | $R+$ | $R-$ | $p$-value | $\alpha = 0.05$ |
|---|---|---|---|---|
| jDE | 1485.5 | 284.5 | **6.00E-06** | **Yes** |
| SaDE | 1765.5 | 4.5 | **0.00E+00** | **Yes** |
| EPSDE | 1372.5 | 397.5 | **2.30E-04** | **Yes** |
| JADE | 1388.0 | 382.0 | **1.44E-04** | **Yes** |
| CoDE | 1657.0 | 173.0 | **0.00E+00** | **Yes** |
| CoBiDE | 1407.0 | 423.0 | **2.88E-04** | **Yes** |
| SinDE | 1384.0 | 386.0 | **1.63E-04** | **Yes** |
| MPEDE | 1275.5 | 494.5 | **3.16E-03** | **Yes** |

Table 13 Comparison results of MLCCDE with start-of-the-art and up-to-date DE variants on 30- and 50-dimensional CEC2017 benchmark set according to multi-problem Wilcoxon's test

| MLCCDE **v.s.** | $R+$ | $R-$ | $p$-value | $\alpha = 0.05$ |
|---|---|---|---|---|
| jDE | 1764.0 | 66.0 | **0.00E+00** | **Yes** |
| SaDE | 1737.0 | 33.0 | **0.00E+00** | **Yes** |
| EPSDE | 1607.5 | 222.5 | **0.00E+00** | **Yes** |
| JADE | 1588.0 | 242.0 | **1.00E-06** | **Yes** |
| CoDE | 1700.5 | 69.5 | **0.00E+00** | **Yes** |
| CoBiDE | 1578.0 | 252.0 | **1.00E-06** | **Yes** |
| SinDE | 1469.5 | 300.5 | **1.00E-05** | **Yes** |
| MPEDE | 1413.0 | 417.0 | **2.43E-04** | **Yes** |

Table 14 Overall performance ranking of all the considered
DEs on 30 and 50-dimensional CEC2014 and cec2017
benchmark set by Friedman's test

| CEC2014 | | CEC2017 | |
|---|---|---|---|
| Algorithm | Ranking | Algorithm | Ranking |
| MLCCDE | **2.78** | MLCCDE | **2.49** |
| MPEDE | 4.21 | MPEDE | 3.86 |
| CoBiDE | 4.46 | CoBiDE | 4.73 |
| JADE | 4.85 | JADE | 4.78 |
| SinDE | 4.99 | SinDE | 4.84 |
| CoDE | 5.06 | CoDE | 5.40 |
| jDE | 5.10 | jDE | 5.76 |
| EPSDE | 6.13 | SaDE | 6.33 |
| SaDE | 7.38 | EPSDE | 6.77 |

## 4.7 Flexibility of MLCC

To further demonstrate the flexibility of the framework, two experiments were designed as follows.

In the first experiment, an example of utilizing MLCC to incorporate three optimizers is presented. The three previously used algorithms, i.e. SHADE, IDE and BiDE are considered. It is noticed that SHADE and BiDE share some similarities as they adopt the same mutation strategy. The pseudocode of MLCC-SIBi is given in Algorithm S-3 in the supplemental file. Parameter settings for the algorithms are set the same as used in Sec. 4.1 and Sec. 4.4. As observed in Table S17 and Table 15, the MLCC variant MLCC-SIBi exhibits better performance compared to the baseline DEs. More specifically, MLCC-SIBi performs better in 40 (=16+15+9) cases and underperforms in 15(=5+4+6) cases on the 30-dimensional functions. For the 50-dimensional case, MLCC-SIBi wins in 56(=20+16+20) functions and loses in 7(=2+4+1) functions.

Table 15 Comparison results of MLCC-SIBi with its baseline
DE variants on 30- and 50-dimensional CEC2014 benchmark
set according to Wilcoxon signed-rank test with a
significance level of 0.05

| -/=/+ | $D = 30$ | $D = 50$ |
|---|---|---|
| SHADE | 16/9/5 | 20/8/2 |
| IDE | 15/11/4 | 16/10/4 |
| BiDE | 9/15/6 | 20/9/1 |

In the second experiment, MLCC was extended to incorporate the L-SHADE [34] algorithm with linear population size reduction (LPSR). To this end, L-SHADE and M_IDE, are assigned to the two layers, respectively. M_IDE is a modified version of IDE with the original parameter strategy replaced by the success history-based parameter adaption (SHA) [33]. The reason for this strategy replacement is that performance of the original parameter strategy in IDE degrades with the LPSR scheme.

The graphic illustration and pseudocode of the resulting MLCC-L-SI variant are shown in Fig. S1 and Algorithm S-4 in the supplemental file, respectively.

Parameter settings for the algorithms are summarized as follows.

1) L-SHADE: $NP^{init} = 20 \times D$, $M_F = \{0.7\}$, $M_{CR} = \{0.5\}$, and $H = 5$.

2) M_IDE: $NP = 5 \times D$, $M_F = \{0.7\}$, $M_{CR} = \{0.5\}$, and $H = NP$.

3) MLCC-L-SI: $NP^{init} = 20 \times D$, $M^{LSHA}_F = \{0.7\}$, $M^{LSHA}_{CR} = \{0.5\}$, $H^{LSHA} = 5$ (For L-SHADE layer), $NP^{M\_IDE} = 5 \times D$, $M^{M\_IDE}_F = \{0.7\}$, $M^{M\_IDE}_{CR} = \{0.5\}$, $H^{M\_IDE} = 5$ (For M_IDE layer), and $N = 0.05$.

*Remark:* In our experiment, M_IDE maintains a fixed population size $NP$ to ensure good performance and the history length $H$ is set to the population size $NP$, as recommend in SHA [33]. While in MLCC-L-SI, the population size of the M_IDE layer $NPT_G$ is fixed at $5 \times D$ when the current population size $NP_G \geq 5 \times D$. However, when $NP_G < 5 \times D$, $NPT_G$ is also adjusted according to the LPSR scheme, as shown in Fig. S1. Thus, the history length $H^{M\_IDE}$ is set the same as $H^{LSHA}$ for simplicity.

As shown in Table S18 and Table 16, MLCC-L-SI exhibits better performance than the constituent algorithms, winning in 60 (=7+19+11+23) cases and losing in 17 (=3+4+6+4) cases. It is also observed that the superiority of MLCC-L-SI over M_IDE is more significant than over L-SHADE. The reason lies in that the performance of M_IDE is significantly inferior to that of L-SHADE, as shown in Table S18. Nevertheless, MLCC-L-SI still achieves better performance compared to L-SHADE.

Table 16 Comparison results of MLCC-L-SI with its baseline
DE variants on 30- and 50-dimensional CEC2014 benchmark
set according to Wilcoxon signed-rank test with a
significance level of 0.05

| -/=/+ | $D = 30$ | $D = 50$ |
|---|---|---|
| L-SHADE | 7/20/3 | 11/13/6 |
| M_IDE | 19/7/4 | 23/3/4 |

## 5. Conclusion

In this paper, a multi-layer competitive-cooperative (MLCC) framework with a new parallel structure is proposed. The framework can effectively incorporate multiple competitive DE variants and combine their advantages. As a result, the design outperforms all of the constituents. MLCC consists of two components, namely the individual preference layer selecting (IPLS) mechanism and the resource allocation bias (RAB) scheme. The IPLS allows bidirectional information communication between population and multiple adaptive optimizers assigned in multiple layers, making the optimizers work in a collaborative manner. The RAB provides an effective resource allocation, to promote the searching capability. The effectiveness and advantages of the MLCC framework as well as its components are confirmed by comprehensive experiments carried out on the CEC benchmark functions.

In this study, mainly two or three DE methods are incorporated into MLCC. We suggest some heuristic ways to demonstrate how these methods are selected. However, it is still open as to how best determine the set of methods. Moreover, it will be interesting to see how the proposed framework can be extended to other EAs, which is another direction for future work.

The MATLAB code of MLCC can be downloaded from https://zsxhomepage.github.io/.

## Acknowledgments

The work was supported in part by the National Natural Science Foundation of China (No. 61671485), in part by the International Science & Technology Cooperation Program of China (No. 2015DFR11050), and in part by City University of Hong Kong under a SRG Grant (Project No: 7004710).